\documentclass{article}

\usepackage[main,final]{nips2026/neurips_2026}

\usepackage[utf8]{inputenc} 
\usepackage[T1]{fontenc}    
\usepackage{hyperref}       
\usepackage{xurl}
\usepackage{booktabs}       
\usepackage{amsfonts}       
\usepackage{nicefrac}       
\usepackage{microtype}      
\usepackage[table]{xcolor}         

\usepackage{multirow}
\usepackage{amssymb}
\usepackage{amsmath}
\usepackage{graphicx}

\usepackage{algorithm}
\usepackage{algpseudocode}

\definecolor{myBlue}{HTML}{DEECF4}  
\definecolor{BrickRed}{RGB}{203, 65, 84}
\definecolor{SoftGreen}{RGB}{51, 153, 51}
\definecolor{SoftGray}{HTML}{929AA2}

\title{RollVerify: Bridging Efficiency and Accuracy in Long-Tail Rollout Reinforcement Learning}

\author{
  \vspace{-25pt}\\
  \textbf{Yongqiang Yao$^{1,\dag}$,\quad Jinru Tan$^{3,\dag}$,\quad Kaihuan Liang$^{2,\dag}$,\quad \quad Zixin Yin$^2$} \\
 \textbf{ \quad Yazhe Niu$^4$,  \quad Ruihao Gong$^{5*}$, \quad Dahua Lin$^2$, \quad Ningyi Xu$^{1}$\thanks{Corresponding authors: \href{mailto:gongruihao@buaa.edu.cn}{\color{black}{gongruihao@buaa.edu.cn}}, \href{mailto:xuningyi@sjtu.edu.cn}{\color{black}{xuningyi@sjtu.edu.cn}}; \, $\dag$: equal contribution } } \\
  $^1$Shanghai Jiao Tong University \quad $^2$SenseTime Research \quad $^3$Central South University  \\
  \quad $^4$The Chinese University of Hong Kong \vspace{3pt} \quad $^5$Beihang University \\
  \texttt{\small soundbupt@gmail.com, \small tanjingru@csu.edu.cn, liangkaihuan@sensetime.com} \\ 
  \texttt{\small {yinzixin}@sensetime.com, niuyazhe314@outlook.com, } \\
  \texttt{\small gongruihao@buaa.edu.cn, dhlin@sensetime.com, xuningyi@sjtu.edu.cn}\vspace{8pt}  \\
  }

\begin{document}

\maketitle

\begin{abstract}


Reinforcement learning is crucial for improving large language models’ reasoning and generalization.  It relies on massive rollouts whose lengths become increasingly long-tailed as context windows grow. In on-policy training, these long-tail rollouts can result in GPU bubbles, reducing system utilization and limiting RL scalability. Asynchronous or partial-rollout methods improve throughput by relaxing synchronization, but inevitably introduce stale off-policy samples (trajectories) that may hurt final accuracy. Existing approaches mainly mitigate this off-policy issue by reweighting off-policy samples during training, yet they can still leave a performance gap compared to fully on-policy training. In this work, rather than \textbf{passively reweighting samples during training}, we propose RollVerify, a lightweight RL framework built on partial rollout that \textbf{actively verifies and repairs samples before they enter training.} Specifically, it introduces an \textbf{o}ff-\textbf{p}olicy \textbf{s}hift metric OPS, to quantify the off-policy deviation of partially generated trajectories. Guided by the OPS constraint, RollVerify performs both sequence-level and token-level verification to identify and truncate invalid suffixes of trajectories. This yields high-quality samples that protect the models’ accuracy while preserving the efficiency gains of partial rollout. Experiments on mathematical and tool-assisted mathematical reasoning show that RollVerify achieves accuracy comparable to on-policy training while reducing training cost. Additional code-generation results provide preliminary evidence beyond mathematics.

\end{abstract}

\section{Introduction}





\begin{figure}[t]
    \centering
    \includegraphics[width=0.95\textwidth]{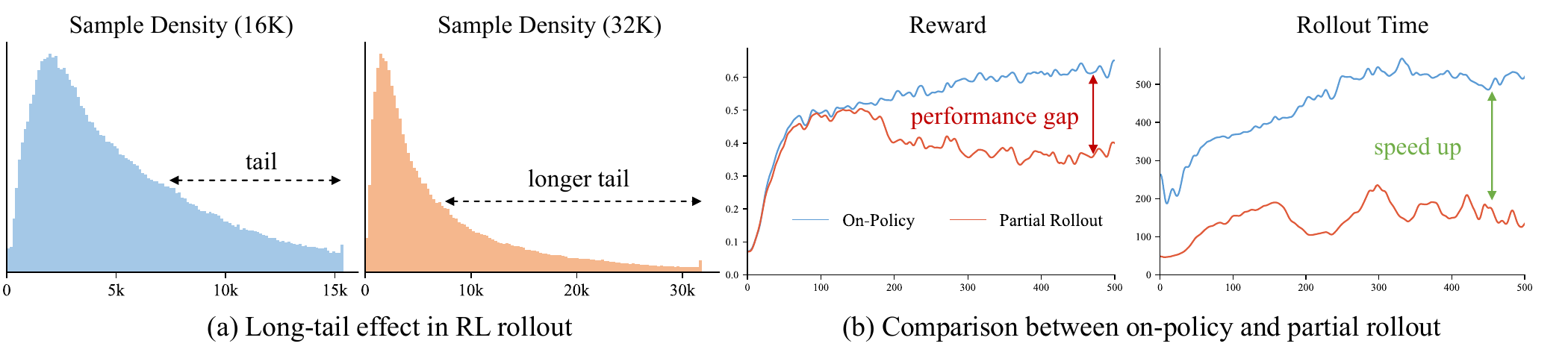}
    \caption{\textbf{(a) Long-tail response length distributions observed} during RL rollout under 16K and 32K settings. \textbf{(b) Comparison between the standard on-policy and partial rollout} in terms of speed and performance.}
    \label{fig:intro}
\end{figure}

Reinforcement learning (RL) has become a crucial paradigm for enhancing the reasoning and decision-making capabilities of large language models~\cite{guo2025deepseekr1, ouyang2022instructgpt,yang2025qwen3technicalreport,grattafiori2024llama3herdmodels,kimiteam2026kimik25visualagentic,glm5team2026glm5vibecodingagentic,anthropic2025claude,google2025gemini3}. In a typical on-policy RL pipeline, the rollout phase, which involves collecting diverse rollouts from the current policy, occupies 70-80\% of the total time~\cite{gao2025rollpacker,fu2025areal}. In practice, it is observed that rollouts often exhibit a long-tail distribution, see Figure~\ref{fig:intro}~(a). Specifically, some samples terminate quickly, whereas others require much longer generation times. The long-tail effect~\cite{fu2025areal, han2025asyncflowasynchronousstreamingrl} causes GPUs to idle while waiting for a few straggling samples, leading to GPU bubbles and reducing overall training efficiency.

One approach mitigates GPU bubbles through system-level optimizations~\cite{xiaomi2025mimo, wang2025roll}, while maintaining the on-policy nature of the training samples (rollout trajectories). However, the inflexible system and limited improvements hinder their practicality. Other methods relax the strict on-policy constraint, exploring asynchronous techniques to enhance efficiency~\cite{deepscaler2025,zhong2025streamrlscalableheterogeneouselastic,han2025asyncflowasynchronousstreamingrl}. These approaches allow training to use rollouts generated by earlier policies, decoupling rollout generation from model training, which reduces synchronization overhead and improves resource utilization.

Partial rollout is a simple variant of asynchronous rollout generation~\cite{slime_github, team2025kimi}.
When short trajectories finish, the freed compute immediately generates the next trajectory instead of waiting for the long trajectories to complete. Once enough training samples are collected, the unfinished long trajectories are paused and will be resumed later.
It reduces GPU bubbles and improves overall throughput, as shown in Fig.~\ref{fig:intro}~(b)-right. However, training samples may be generated by stale policies, introducing off-policy trajectories that may degrade final accuracy, as shown in Fig.~\ref{fig:intro}~(b)-left.

Existing methods mainly mitigate the impact of off-policy samples at the loss level by reweighting their contribution during training, such as GSPO~\cite{zheng2025gspo}, SAPO~\cite{gao2025sapo}, and VESPO~\cite{shen2026vespo}. However, such loss-level corrections only reshape the training gradients after trajectories have been generated; they merely try to tolerate off-policy trajectories but not modify them. Highly off-policy samples still remain in the training data and may hurt final accuracy.

To address this limitation, in this work, we propose \textbf{RollVerify} to ensure each generated sample for training is \textit{nearly on-policy}. It builds on partial rollout, but adds a lightweight verification phase to actively verify and repair trajectories before they enter training. We first design a novel metric, \textit{off-policy shift} (OPS), in Section~\ref{sec:OPS}, to quantify the off-policy degree of generated samples, and validate its correlation with final accuracy. For unfinished trajectories, we applied an OPS-based two-stage verification method in Section~\ref{sec:method_verification} that detects off-policy shift at both the sequence and token levels. Highly off-policy parts are truncated and regenerated to repair the trajectory. RollVerify preserves most of the throughput benefit of partial rollout by avoiding GPU bubbles, while substantially eliminating the accuracy degradation caused by stale or mixed-policy trajectories. Therefore, it provides a better efficiency--accuracy trade-off for long-tail rollout RL training.

We evaluate our method across diverse models, scales, and task domains. On DAPO-Math~\cite{yu2025dapo} with Qwen3-8B~\cite{yang2025qwen3technicalreport}, RollVerify reduces training cost from 98.1 to 57.5 GPU days while preserving accuracy. RollVerify achieves comparable accuracy to on-policy training with lower training cost on both dense and MoE models. Additional experiments cover tool-assisted mathematical reasoning and provide preliminary evidence on code generation. Our contributions are threefold. First, we design a new metric to quantify the off-policy degree of generated samples. Second, we propose RollVerify, which mitigates the impact of off-policy samples through rollout-level quality control, achieving a better balance between training efficiency and accuracy. Third, our experiments demonstrate RollVerify's effectiveness across the evaluated models and task settings.

\section{Related Works}

\noindent \textbf{Reinforcement Learning for LLMs.} Reinforcement learning (RL) has become a core component of LLM post-training, particularly for enhancing complex reasoning capabilities~\cite{guo2025deepseekr1}. Proximal Policy Optimization (PPO)~\cite{ppo,ouyang2022instructgpt} was among the earliest methods adopted in LLM post-training, serving as a standard foundation for subsequent RLHF and reasoning-oriented pipelines. More recently, Group Relative Policy Optimization (GRPO)~\cite{grpo}, along with several of its improved variants~\cite{yu2025dapo,gao2025sapo,zheng2025gspo,minimax2025minimaxm1scalingtesttimecompute}, has emerged as an effective alternative that avoids explicit value modeling while retaining strong empirical performance. Nevertheless, these methods still rely on large amounts of rollout trajectories collected from the current or recent policy. Therefore, rollout quality, diversity, and system throughput are critical to scalable RL for LLMs.

\noindent \textbf{RL Training Acceleration.} Large-scale rollout has become a major bottleneck in LLM RL training. To improve rollout efficiency, high-throughput serving systems such as vLLM~\cite{vllm} and SGLang~\cite{zheng2024sglang} are widely adopted in practice. More recent studies~\cite{fu2025areal,han2025asyncflowasynchronousstreamingrl} have further examined the long-tail effect in rollout generation, where large variance in sample lengths leads to workload imbalance and GPU idle. Existing approaches can be broadly divided into two categories. The first category preserves strict on-policy training and mitigates the long-tail effect through system-level optimizations, such as overlapping rollout and reward computation~\cite{wang2025roll,xiaomi2025mimo,zhong2024rlhfuseefficientrlhftraining} or improving rollout scheduling~\cite{gao2025rollpacker}. These methods typically rely on specialized pipeline designs, while offering only limited improvements in flexibility and efficiency. The second category relaxes the on-policy constraint through asynchronous designs~\cite{fu2025areal, han2025asyncflowasynchronousstreamingrl} to improve resource utilization and training throughput. However, such methods inevitably introduce off-policy samples, which may degrade model accuracy. In contrast, our method places greater emphasis on \textit{preserving model accuracy while retaining the efficiency benefits} of asynchronous rollout.

\noindent \textbf{Off-Policy Effect Handling.} Handling off-policy data has long been a fundamental challenge in policy optimization. PPO mitigates policy mismatch through importance weighting and clipping. Recent studies mitigate off-policy effects mainly at the loss level by reweighting the contribution of off-policy samples during training. GSPO~\cite{zheng2025gspo} reduces off-policy effects through sequence-level optimization. SAPO~\cite{gao2025sapo} replaces hard clipping with soft adaptive weighting to provide a smoother correction. VESPO~\cite{shen2026vespo} corrects sequence-level importance weights by introducing a variational reshaping kernel. However, these methods only adjust the contribution of off-policy samples \textit{during training}. Low-quality samples may still be retained and may affect final model accuracy. In contrast, our method intervenes earlier in the pipeline: instead of only reweighting off-policy samples after generation, we identify and repair partially generated trajectories \textit{before they enter training}, therefore producing higher-quality training samples.


\section{Problem Analysis}

\subsection{Preliminary}

\noindent \textbf{Group Relative Policy Optimization (GRPO).} Given a prompt $x$, a group of responses $\{y^{(i)}\}_{i=1}^{G}$ is sampled, and the reward $R$ is computed for each response.
Let $\pi_\theta$ be the current policy and $\pi_{\mathrm{old}}$ the behavior policy used for rollout collection. GRPO optimizes the clipped surrogate objective:

\begin{equation}
\mathcal{L}_{\mathrm{GRPO}}(\theta)
=
\mathbb{E}_{t}\!\left[
\min\!\left(
r_t(\theta)\hat{A}_t,\;
\mathrm{clip}\!\left(r_t(\theta),1-\epsilon,1+\epsilon\right)\hat{A}_t
\right)
\right]
\label{eq:ppo_loss}
\end{equation}


where $\hat{A}_t$ denotes the advantage estimate and $r_t(\theta)$ is the importance ratio that corrects the mismatch between the current policy and the behavior policy, computed following Eq.~\ref{eq:grpo_ada}. The clipping operator $\mathrm{clip}\!\left(r_t(\theta), 1-\epsilon, 1+\epsilon\right)$ further restricts the update when $r_t(\theta)$ deviates too much from $1$.
\begin{equation}
\hat{A}^{(i)}_{t} 
=
\frac{R^{(i)}- \mathrm{mean}(\{R^{(j)}\}_{j=1}^G)}
{\mathrm{std}(\{R^{(j)}\}_{j=1}^G)},
\quad
r_t(\theta)=
\frac{\pi_\theta(y_t \mid x,y_{<t})}
{\pi_{\mathrm{old}}(y_t \mid x,y_{<t})}
\label{eq:grpo_ada}
\end{equation}

\noindent \textbf{Partial Rollout.} Partial rollout~\cite{team2025kimi} follows a repeated rollout--training cycle. In the rollout phase, multiple rollout engines generate trajectories in parallel, and engines that finish shorter trajectories early are immediately reassigned to new rollout tasks. Once enough completed samples are collected to form a training batch, the training phase starts. Unfinished long trajectories are paused and resumed in later rollout phases rather than discarded.



\subsection{Measuring Off-Policy Deviation}
\label{sec:OPS}

We define \textbf{Off-Policy Shift (OPS)} to quantify the degree to which a trajectory deviates from the current policy. In Eq \ref{eq:OPS}, for each token $y_t$ at trajectory position $t$, the token-level OPS $D(y_t)$ is defined as the absolute offset of the importance ratio between the current policy $\pi_\theta$ and the behavior policy $\pi_{\mathrm{old}}$ from $1$. We then define the sequence-level off-policy shift $D(\tau)$ as the average token-level $D(y_t)$ over the whole trajectory $\tau$:

\begin{equation}
D(y_t) = \left| \frac{\pi_\theta(y_t \mid x, y_{<t})}{\pi_{\mathrm{old}}(y_t \mid x, y_{<t})} - 1 \right|,
\quad
D(\tau) = \frac{1}{T}\sum_{t=1}^{T} D(y_t).
\label{eq:OPS}
\end{equation}

Intuitively, $D(\tau)=0$ indicates an on-policy trajectory, where the importance ratio equals $1$ at each position $t$. A larger shift $D(\tau)$ implies a higher degree of off-policy for this trajectory $\tau$.

\begin{table}[t]
\centering
\small
\setlength{\abovecaptionskip}{2pt}
\setlength{\tabcolsep}{4pt}
\renewcommand{\arraystretch}{1.15}
\begin{tabular}{l|lc|cccc}
\toprule
\textbf{Setting} & \textbf{Filter} & \textbf{Threshold}  & \textbf{AIME 24} $\uparrow$ & \textbf{AIME 25} $\uparrow$ & \textbf{OPS} & \textbf{GPU days}\\
\midrule
GRPO baseline (on-policy) & None & -- & \underline{29.4} & \textbf{27.4} & 0 & 30.1~(1.00x) \\
\midrule
\multirow{7}{*}{Partial Rollout (off-policy)}
& None      & --    & 19.5 & 19.1 & 0.025 & 19.4~(\textcolor{SoftGreen}{1.55x})\\
& Staleness & 1     & 21.2 & 20.9 & 0.017 &  26.6~(\textcolor{SoftGray}{1.13x})\\
& Staleness & 2     & 19.7   & 19.3   & 0.018 & 21.5~(\textcolor{SoftGreen}{1.40x})\\
& Staleness & 3     & 18.7 & 18.4 & 0.018 & 19.8~(\textcolor{SoftGreen}{1.52x})\\
& OPS       & 0.02  & 21.1   & 20.3   & 0.016    & 27.1~(\textcolor{SoftGray}{1.11x})\\
& OPS       & 0.01  & 27.7 & 24.2 & 0.009 & 29.0~(\textcolor{SoftGray}{1.04x})\\
 & OPS       & 0.005  & \textbf{29.9}   & \underline{26.4}   & 0.004 & 33.3~(\textcolor{BrickRed}{0.90x}) \\
\bottomrule
\end{tabular}
\caption{Analysis results of staleness-based and OPS-based filtering strategies under partial rollout on DAPO-Math. The baseline setting uses Qwen3-4B-Base, with a batch size of 128. }
\label{tab:problem_analysis}
\end{table}
\begin{figure}[t]
    \centering
    \includegraphics[width=0.95\textwidth]{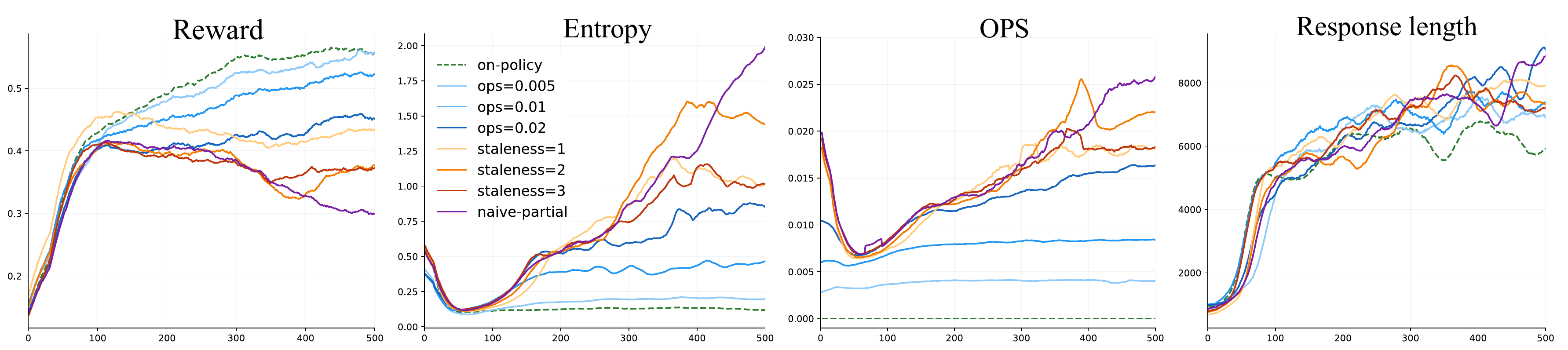}
    \caption{Training dynamics under different partial rollout settings using Qwen3-4B-Base model. We train the model for 500 steps in total, with the x-axis denoting the training step. }
    \label{fig:analysis}
\end{figure}

\subsection{Sample Filtering Analysis}
\label{sec:sample_filtering_analysis}

To validate the effectiveness of the proposed metric, we conduct off-policy sample filtering experiments based on OPS and compare them with an existing staleness-based filtering strategy. In both settings, samples that do not satisfy the filtering criterion are directly discarded.

\noindent \textbf{Staleness-based Filtering.}
\cite{team2025kimi} adopts a simple \textit{staleness-based} strategy that discards partial rollout samples whose staleness exceeds a predefined threshold. This heuristic assumes that older samples (larger staleness) are more likely to destabilize training. However, our results show that staleness is only a coarse proxy for off-policy deviation. As shown in Table~\ref{tab:problem_analysis} (rows 3--5), and Fig.~\ref{fig:analysis}, staleness filtering still leads to entropy explosion, causing accuracy degradation. We also observed that its OPS remains high during training.


\noindent \textbf{OPS-based Filtering.} We further study an alternative strategy based on our proposed OPS. 
Specifically, given a predefined threshold $\gamma$, we directly discard samples whose sequence-level OPS $D(\tau)$ exceeds the threshold. As shown in Table~\ref{tab:problem_analysis} (rows 6-8), this strategy substantially improves model accuracy and stability, which can be further verified in Fig.\ref{fig:analysis}.
Moreover, stricter thresholds generally lead to larger gains, showing a clear positive correlation between performance and off-policy control. However, a strict threshold discards a large number of samples, which significantly reduces the effective training throughput.

\section{RollVerify}

\noindent \textbf{Motivation.} The analysis in Section~\ref{sec:sample_filtering_analysis} shows a clear trade-off between accuracy and efficiency. 
On the one hand, OPS-based filtering can strictly control the off-policy degree of rollout samples and protect model accuracy. On the other hand, directly discarding samples wastes computation and reduces effective training throughput. Our purpose is to preserve accuracy without sacrificing the efficiency advantage of partial rollout training.

Inspired by the \textbf{verify-and-truncate principle} in multi-token prediction (MTP)~\cite{liu2024deepseekv3} and speculative decoding~\cite{leviathan2023fastinferencetransformersspeculative}: candidate tokens are not rejected as a whole; rather, the verified prefix is accepted, while the suffix after the first verification failure is discarded. Similarly, when a rollout violates the OPS requirement, we can retain the prefix that still satisfies OPS and remove only the invalid suffix. This strategy avoids wasting partially valid rollouts and improves sample utilization, while preventing invalid off-policy segments from destabilizing training.



\subsection{Overall Procedure}
\label{sec:method_overall}

\begin{figure}[t]
    \centering
    \includegraphics[width=1.0\textwidth]{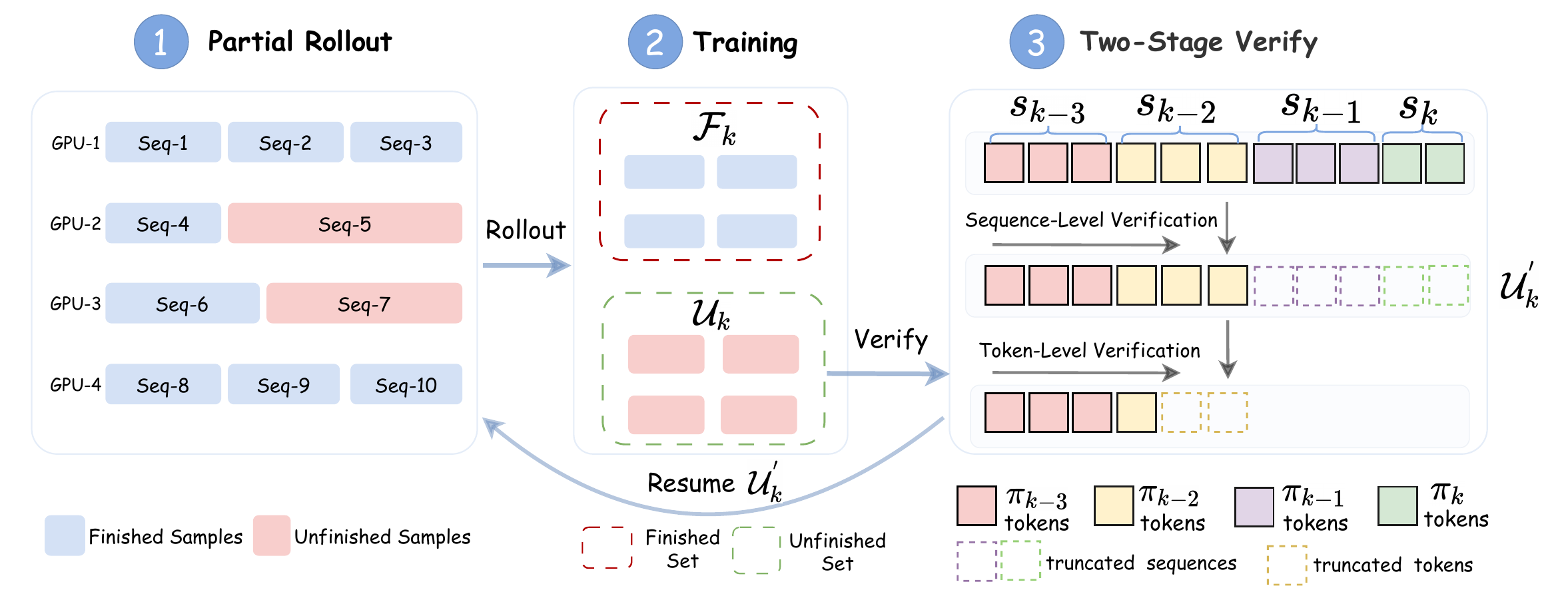}
    \caption{\textbf{Overview of RollVerify.} It extends the standard rollout--training loop with an additional verification phase. \textbf{left:} rollout phase; \textbf{mid:} training phase; \textbf{right:} proposed verification phase.}
    \label{fig:method}
\end{figure}

Figure~\ref{fig:method} illustrates the overall framework of RollVerify. We consider a colocated architecture in which rollout and training share the same GPU pool and execute in alternating phases. During rollout, freed resources are continuously assigned to new trajectories. After training, the updated parameters are synchronized to the rollout workers before the next rollout phase. This setting differs from fully asynchronous, disaggregated architectures, where rollout and training execute concurrently on separate GPU pools. Within this colocated setting, RollVerify extends the rollout–training loop with an additional verification phase. We describe each phase in detail below.

\textbf{Rollout Phase.} At training step $k$, the current policy is denoted as $\pi_k$.
The partial rollout keeps generating samples using $\pi_k$, and completed samples are added to the finished set $\mathcal{F}_k$.
Once the size of finished set $|\mathcal{F}_k|$ reaches the target training batch size, the rollout phase stops and the partially generated samples are cached in the unfinished set $\mathcal{U}_k$. After rollout, we move to training phase.

\textbf{Training Phase.} The samples in $\mathcal{F}_k$ are directly used for \textit{one optimization step}, which updates the policy from $\pi_k$ to a new policy $\pi_{k+1}$. At this point, the unfinished samples in $\mathcal{U}_k$ become stale with respect to $\pi_{k+1}$, since they are generated under old policies $\pi_k$, even older policies $\pi_{k-1}, \pi_{k-2}, \pi_{k-3}$. After training, we move to verification phase.



\textbf{Verification Phase.} We apply a two-stage OPS-based verification procedure (in Section~\ref{sec:method_verification}) to the unfinished set $\mathcal{U}_k$ at both the sequence and token levels.
For each unfinished sample, we truncate the highly off-policy suffix and retain only the verified prefix. The verified unfinished set $\mathcal{U}'_k$, which contains both truncated samples and samples that pass verification without truncation, will be resumed in the next rollout phase under the updated policy $\pi_{k+1}$.
The overhead of verification is lightweight; more details are shown in Table~\ref{tab:overhead}.


This process is repeated across training steps. As a result, a sample may contain multiple segments that come from different policies $, \pi_{k-1}, \dots$. Since any partially generated sample is verified before it is resumed, we ensure that all samples used for the next training phase remain \textit{nearly on-policy} with respect to the latest policy. In this way, RollVerify maintains comparable performance to on-policy training while preserving the efficiency benefits of partial rollout.

\begin{figure}[t]
\centering

\begin{minipage}[t]{0.48\linewidth}
\vspace{0pt}
\begin{algorithm}[H]
\caption{OPS Computation}

\small
\begin{algorithmic}[1]
\Require $\tau_{in}=(s_1,\ldots,s_{k-1})$,  $\pi_k$
\Statex \hspace{\algorithmicindent} $p=(p_1,\ldots,p_{k-1})$
\Ensure $\tau_{out}$, $D$

\State $s_k \gets \mathrm{Rollout}(\tau_{in}; \pi_k)$

\State $\tau_{out} \gets(s_1,\ldots,s_{k-1},s_{k})$
\State $p_k \gets \mathrm{ComputeProb}(\tau_{\mathrm{out}}; \pi_k)[k]$
\State $p^{\rm old}\gets(p_1,\ldots,p_{k-1},p_{k})$
\State $\pi_{k+1} \gets \mathrm{UpdateWeights}(\pi_k)$
\State $p^{\rm new} \gets \mathrm{ComputeProb}(\tau_{\mathrm{out}}; \pi_{k+1})$

\State

\For{$i=1$ to $|\tau_{out}|$}
    \For{$j=1$ to $|s_i|$}
        \State $D_{i,j}\gets\left|p^{\rm new}_{i,j}/p^{\rm old}_{i,j}-1\right|$
    \EndFor
\EndFor
\State
\State \Return $\tau_{\rm out},D$

\end{algorithmic}
\label{alg:OPS_compute}
\end{algorithm}
\end{minipage}
\hfill
\begin{minipage}[t]{0.48\linewidth}
\vspace{0pt}
\begin{algorithm}[H]
\caption{Two-Stage Verification}
\label{alg:two_stage_verify}
\small
\begin{algorithmic}[1]
\Require $\tau_{\rm out}$, $D$, $d_s,d_t$, $m \gets 0$
\Ensure $T_{\rm out}$

\For{$i=1$ to $|\tau_{\rm out}|$}
    \State $\bar D_i\gets \sum_j D_{i,j} / |s_i|$
    \If{$\bar D_i\ge d_s$} \textbf{break} \EndIf
    \State $m \gets i$
\EndFor

\State \textbf{if} $m = 0$ \textbf{then} \Return $\emptyset$
\State $T_{\rm out}\gets \rm{concat}(s_1,\dots, s_m), n \gets 0$
\For{$i=1$ to $m$, $j=1$ to $|s_i|$}
    \If{$D_{i,j}\ge d_t$}
        \State \Return $T_{\rm out}[:n]$
    \EndIf
    \State $n\gets n+1$
\EndFor
\State \Return $T_{\rm out}$
\end{algorithmic}
\end{algorithm}
\end{minipage}

\end{figure}

\subsection{Two-Stage Verification}
\label{sec:method_verification}

We first perform OPS computation on the partial trajectory as shown in Algorithm~\ref{alg:OPS_compute}. Then, we conduct two-stage verification (in Algorithm~\ref{alg:two_stage_verify}), including sequence-level and token-level verification.

\noindent \textbf{OPS Computation.} At training step $k$ with current policy $\pi_k$, consider a partial trajectory \(\tau_{\mathrm{in}}=(s_1,\ldots,s_{k-1})\) in unfinished set  $\mathcal{U}_k$, whose segment $s_i$ is generated by the corresponding policy $\pi_i$. The segment probabilities \( p = (p_1,\ldots,p_{k-1}) \) have also been cached, where each \(p_i\) denotes the probability of segment \(s_i\) under its generating policy $\pi_i$. In rollout phase, we resume this trajectory \(\tau_{\mathrm{in}}=(s_1,\ldots,s_{k-1})\) and generate a new segment \(s_k\) using current policy $\pi_k$, resulting in the extended trajectory \(\tau_{\mathrm{out}}=(s_1,\ldots,s_k)\). The new segment probability \(p_k\) is computed under \(\pi_k\), yielding \(p^{\mathrm{old}}=(p_1,\ldots,p_{k-1},p_k)\). In training phase, we optimize the policy from \(\pi_k\) to \(\pi_{k+1}\) and compute the probabilities \(p^{\mathrm{new}}\) of all segments in \(\tau_{\mathrm{out}}\) under the updated policy \(\pi_{k+1}\). The token-level OPS is then computed as \(D_{i,j}=\left|p^{\mathrm{new}}_{i,j}/p^{\mathrm{old}}_{i,j}-1\right|\), where \(i\) indexes the segment in $\tau_{out}$ and \(j\) indexes the token in segment $s_i$.

\noindent \textbf{Sequence-Level Verification.} We first compute average OPS score $\bar D_i$ for each segment $s_i \in \tau_{out}$. It then scans the segments in order from $s_1$ to $s_{k}$. Once it encounters the first segment whose score exceeds the threshold, i.e., $\bar D_i \ge d_s$, it truncates the sample at $s_i$, and we get the 
 verified segments $\{s_{1}, s_{2}, \dots, s_{m}\}$.

\noindent \textbf{Token-Level Verification.} After sequence-level verification, we concatenate all retained segments to a token list  $T_{\rm out}$. We perform token-level verification over  $T_{\rm out}$. We scan each token sequentially and compare its OPS score $D_{i,j}$ with the token threshold $d_t$. If $D_{i,j} \ge d_t$, the procedure stops and returns the verified prefix $T_{\rm out}[:n]$. Otherwise, we accept this token and continue scanning.

\subsection{Conditional Switching}

\noindent \textbf{Acceptance Rate.} The acceptance rate $\alpha_k$ measures the fraction of partially generated tokens retained after verification.
\begin{equation}
\alpha_k = \frac{N^{\mathrm{keep}}}{N^{\mathrm{all}}},
\end{equation}
where $N^{\mathrm{keep}}$ is the number of tokens that are kept after verification, and $N^{\mathrm{all}}$ is the total number of tokens of the unfinished set during the rollout phase.

When samples generated by stale policies become highly off-policy, most of them are directly discarded during verification, which lowers effective resource utilization, making partial rollout inefficient. To avoid this inefficiency, we adopt a conditional switching strategy, which switches from partial rollout to fully on-policy training conditioned on the acceptance rate.  Specifically, we monitor the average acceptance rate over a recent time window. If this average falls below a predefined threshold, e.g., 0.5, we terminate partial rollout and switch to fully on-policy training for the remaining steps. The average acceptance rate can be estimated using either a sliding window or an exponential moving average (EMA). In our implementation, we use a sliding window of size 8.

\section{Experiments}

\subsection{Experimental Setup}

\textbf{Implementation Details.} We implement all methods based on the VeRL~\cite{sheng2024hybridflow} codebase, utilizing the Qwen3-8B-Base model~\cite{yang2025qwen3technicalreport} by default and the GRPO~\cite{grpo} for policy optimization. All methods use the same colocated architecture, in which rollout and training share the same GPU pool and execute in alternating phases. We adopt decoupled clipping~\cite{yu2025dapo} with $\epsilon_{\text{high}} = 0.3$ and $\epsilon_{\text{low}} = 0.2$. 
We further use truncated importance sampling~\cite{liu-li-2025-rl-collapse} to correct the mismatch between the inference engine vLLM~\cite{vllm} and training engine FSDP~\cite{zhao2023pytorchfsdpexperiencesscaling}, with truncation threshold set to $2$. 
For RollVerify, we set the sequence-level threshold to $d_s = 0.01$, the token-level threshold to $d_t = 2$. Unless otherwise specified, all experiments are conducted on 32 H800 GPUs with maximum context length of 32K, a learning rate of 1e-6, a batch size of 128 with rollout-n=8. Models are trained for up to 500 iterations. For MoE models~\cite{yang2025qwen3technicalreport}, we use routing replay~\cite{ma2025stabilizingmoereinforcementlearning} to stabilize training.

\textbf{Dataset and Evaluation Protocol.} We mainly evaluate our method on mathematical reasoning tasks. 
We use DAPO-MATH~\cite{yu2025dapo} as the training set, and evaluate on AIME24, AIME25, AMC23, and MATH500~\cite{hendrycksmath2021}. 
For each setting, we report the accuracy on each benchmark, the average accuracy across all benchmarks, and the total training time (GPU days).

\begin{table}[t]
\centering
\small
\setlength{\tabcolsep}{4pt}
\setlength{\abovecaptionskip}{2pt}
\renewcommand{\arraystretch}{1.15}
\resizebox{0.99\textwidth}{!}{
\begin{tabular}{l|l|cccclrl}
\toprule
\textbf{Model } & \textbf{Setting} & \textbf{AIME24} & \textbf{AIME25} & \textbf{AMC23} & \textbf{MATH500} & \textbf{AVG} & \textbf{OPS} & \textbf{GPU days} \\
\midrule
\multirow{3}{*}{Qwen3-8B-Base}
& On-policy     & 40.3 & 29.5 & 81.4 & 78.0 & 57.3 & 0&98.1(1.0x) \\
& Partial & 23.7 & 19.4 & 67.5 & 73.2 & 45.9\textcolor{BrickRed}{$\downarrow$} & 0.017&51.6(\textcolor{SoftGreen}{1.9x}) \\
& RollVerify    & 39.8 & 30.5 & 81.5 & 77.2 & 57.2 &0.004 &57.5(\textcolor{SoftGreen}{1.7x}) \\
\midrule
\multirow{3}{*}{Qwen3-30B-A3B-Base}
& On-policy     & 52.3 & 37.8 & 89.7 & 70.3 & 62.5 & 0 &112 (1.0x) \\
& Partial & 36.8 & 28.5 & 77.8 & 69.4 & 53.1\textcolor{BrickRed}{$\downarrow$} & 0.016 &56.1(\textcolor{SoftGreen}{2.0x}) \\
& RollVerify    & 52.0 & 38.0 & 89.8 & 70.5 & 62.5 &0.006 &65.1(\textcolor{SoftGreen}{1.7x}) \\
\bottomrule
\end{tabular}}
\caption{Comparison among on-policy training, partial rollout, and proposed RollVerify.}
\label{tab:main_results}
\end{table}
\begin{figure}[t]
    \centering
    \includegraphics[width=0.95\textwidth]{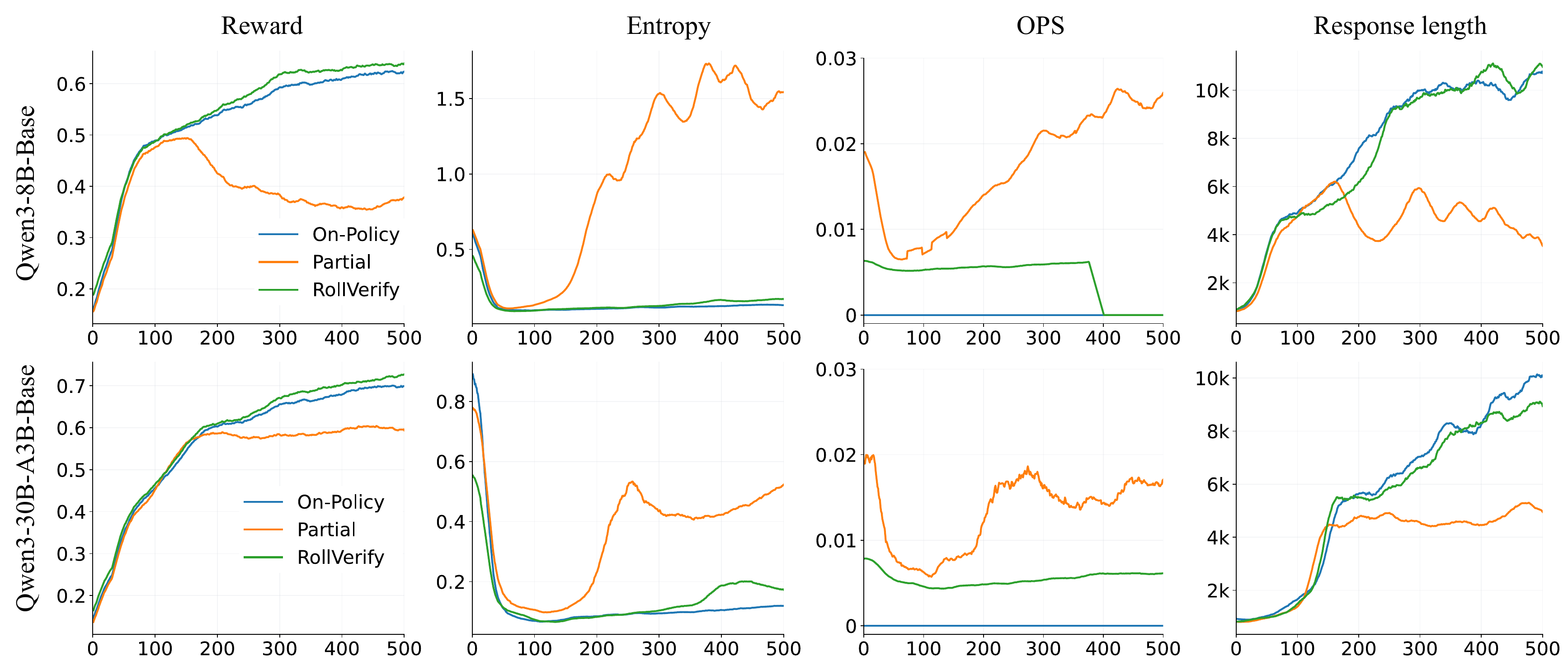}
    \caption{Comparison of training dynamics among On-Policy baseline, Partial, and RollVerify. The first row shows results on Qwen3-8B-Base, and the second row presents results on the MoE model Qwen3-30B-A3B-Base.}
    \label{fig:main_results}
\end{figure}

\subsection{Main Results}

Table~\ref{tab:main_results} compares RollVerify with standard on-policy GRPO and a partial rollout baseline. On Qwen3-8B-Base, our method achieves an average score of 57.2, \textbf{matching the on-policy GRPO baseline} (57.3) while reducing training cost from 98.1 GPU days to 57.5 GPU days, corresponding to a 1.7$\times$ speedup. In contrast, the partial rollout baseline suffers a severe performance drop, with average accuracy decreasing to 45.9, even though it delivers more efficiency gains. Our method maintains on-policy accuracy while achieving a significant speedup.

\noindent \textbf{Results on MoE}. A similar trend is observed on the larger Qwen3-30B-A3B-Base~\cite{yang2025qwen3technicalreport} MoE model, which has more parameters and serves as a stronger baseline than Qwen3-8B-Base. We can see in Table~\ref{tab:main_results} that its on-policy performance is significantly higher, improving the average score from 57.3 (8B dense) to 62.5. Under this stronger and more complex architecture, partial rollout leads to a substantial accuracy degradation. RollVerify is still able to recover the lost performance, matching the on-policy baseline while retaining a 1.7$\times$ training speedup. These results show that RollVerify consistently maintains competitive results across both dense and MoE architectures, demonstrating strong generalization to different models.

\noindent \textbf{Training Stability.} Figure~\ref{fig:main_results} shows that RollVerify preserves training stability and closely matches standard on-policy optimization for both dense and MoE models. Throughout the training dynamics, both entropy and reward remain stable without any collapse. In contrast, naive partial rollout becomes increasingly unstable: entropy rises rapidly and eventually explodes, accompanied by clear drops in both reward and response length. Moreover, RollVerify maintains consistently lower OPS, indicating minimal off-policy degree, whereas naive partial rollout exhibits increasing OPS over time. These results suggest that RollVerify effectively controls off-policy deviation and matchs the stable optimization behavior of on-policy RL.

\subsection{Ablation Study}

\begin{table}[t]
\centering
\small
\setlength{\tabcolsep}{4pt}
\setlength{\abovecaptionskip}{2pt}
\renewcommand{\arraystretch}{1.15}
\resizebox{0.99\textwidth}{!}{
\begin{tabular}{lcccccccc}
\toprule
\textbf{Setting} & \textbf{AIME24} & \textbf{AIME25} & \textbf{AMC23} & \textbf{MATH500} & \textbf{AVG} & \textbf{OPS} &\textbf{AC Rate} &\textbf{GPU Days} \\
\midrule
Naive partial       & 23.7 & 19.4 & 67.5 & 73.2 & 45.9 & \textcolor{BrickRed}{0.017}&1.0 & 51.6  \\
\midrule
+seq-level verify   & 32.6 & 26.5 & 78.5 & 78.1 & 53.9 & 0.009&0.67 & 68.1 \\
+token-level verify & 39.6 & 30.8 & 81.8 & 77.2 & 57.4 &0.005 &0.83 & 61.5 \\
+switch             & 39.8 & 30.5 & 81.5 & 77.2 & 57.2 &\textbf{\textcolor{SoftGreen}{0.004}} & \textbf{0.83} & \textbf{57.5} \\
\bottomrule
\end{tabular}
}
\caption{\textbf{Ablation study of RollVerify under Qwen3-8B-Base setting.} We progressively add sequence-level verification, token-level verification, and conditional switching on top of the naive partial rollout baseline. AC Rate indicates acceptance rate.}
\label{tab:ablation_component}
\end{table}
\begin{figure}[t]
    \centering
    \includegraphics[width=0.98\textwidth]{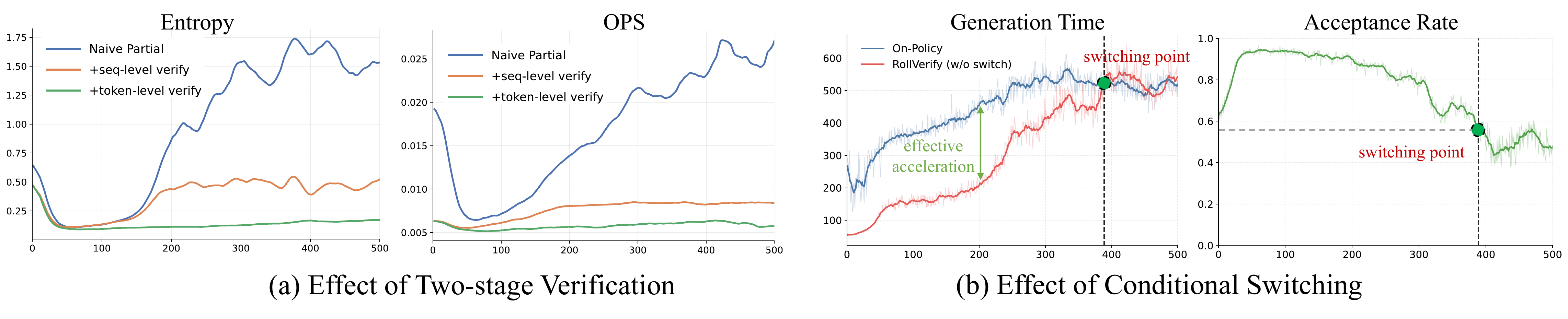}
    \caption{\textbf{Ablation Results.} (a) Effect of Two-stage verification with respect to entropy and OPS. (b) Analysis of conditional switching on generation time and acceptance rate. The dashed vertical line indicates the switching point from partial rollout to fully on-policy training.}
    \label{fig:exp_swtiching}
\end{figure}







\begin{table}[t]
\centering
\begin{minipage}[t]{0.48\linewidth}
\centering
\small
\setlength{\tabcolsep}{4pt}
\renewcommand{\arraystretch}{1.15}
\resizebox{0.99\textwidth}{!}{
\begin{tabular}{l|cc|c|c}
\toprule
\textbf{$d_s$} & \textbf{AIME24} & \textbf{AIME25} & \textbf{AC Rate} & \textbf{GPU Days} \\
\midrule
$\infty$ &23.7& 19.4& 1.0& 51.6 \\
\midrule
0.02  & 28.3 & 24.5 & 0.91 & 53.2 \\
0.015  & 30.4 & 25.2 & 0.80 & 60.2\\
0.012  & 32.4 & 26.7 & 0.75 & 68.0 \\
\rowcolor{myBlue!60}
0.01 & 32.6 & 26.5 & 0.67 & 68.1 \\
0.005 & 39.6 & 30.2 & 0.20 & 123.2 \\
\bottomrule
\end{tabular}
}
\caption{\textbf{Comparison of different $d_s$}. Only sequence-level verification is used.}
\label{tab:hyper_seq}
\end{minipage}
\hfill
\begin{minipage}[t]{0.48\linewidth}
\centering
\small
\setlength{\tabcolsep}{4pt}
\renewcommand{\arraystretch}{1.15}
\resizebox{0.99\textwidth}{!}{
\begin{tabular}{l|cc|c|c}
\toprule
\textbf{$d_t$}  & \textbf{AIME24} & \textbf{AIME25} & \textbf{AC Rate} & \textbf{GPU Days} \\
\midrule
$\infty$ & 32.6 & 26.5 & 0.67 & 68.1 \\
\midrule
1.5  & 40.2 & 30.2 & 0.60 & 70.2 \\
\rowcolor{myBlue!60}
2 & 39.8 & 30.5 & 0.83 & 57.5 \\
3    & 38.5 & 30.4 & 0.84 & 57.2 \\
4    & 38.1 & 30.1 & 0.88  & 55.4\\
5    & 38.0 & 30.2 & 0.89 & 55.2 \\
\bottomrule
\end{tabular}}
\caption{\textbf{Comparison of different $d_t$}. Token and Seq-level verification ($d_s=0.01$) are enabled.}
\label{tab:hyper_token}
\end{minipage}
\end{table}


\noindent \textbf{Importance of Two-Stage Verification.} As shown in Table~\ref{tab:ablation_component}, adding sequence-level verification already leads to a large improvement in model accuracy, raising the average score from 45.9 to 53.9. This shows that removing overly off-policy segments is highly effective in protecting model accuracy. Adding token-level verification further improves the average score from 53.9 to 57.4. Interestingly, it also improves training efficiency, reducing GPU days from 68.1 to 61.5. This is because sequence-level verification only ensures the overall OPS of the sequence but cannot exclude outlier tokens. These highly off-policy tokens affect model stability, causing an increase in OPS and a decrease in acceptance rate, see Figure~\ref{fig:exp_swtiching}~(a). Token-level verification can protect the model from highly off-policy tokens, thus achieving a lower OPS and higher acceptance rate, as shown in Table~\ref{tab:ablation_component}. These results demonstrate the importance of token-level verification.


\noindent \textbf{Analysis of Conditional Switching.} Figure~\ref{fig:exp_swtiching}~(b) shows the generation time and acceptance rate throughout the training process (using our method but without conditional switching). Before the dashed vertical line ($\sim$ iter 400), our method achieves a significantly lower generation time than the GRPO baseline, indicating effective acceleration. However, in the later stage of training, more samples become highly off-policy, causing the acceptance rate to start decreasing dramatically. We can see that the acceleration benefit vanishes when the acceptance rate drops to about $0.5$. At this point, switching back to fully on-policy training can improve the overall training efficiency. As shown in Table~\ref{tab:ablation_component}, adding conditional switching can reduce GPU days from 61.5 to 57.5. At a 16K context length and 700 training steps, RollVerify without switching achieves comparable AIME24/25 accuracy to on-policy training at lower cost, while switching provides additional cost savings (Appendix ~\ref{appendix:no_swtiching}).


\noindent \textbf{Hyper-parameter Analysis.} In Table~\ref{tab:hyper_seq}, we show results using only sequence-level verification under different $d_s$. When $d_s=\infty$, it degenerates to naive partial rollout. A very small $d_s=0.005$~(row6) greatly improves performance but sharply reduces the acceptance rate to 0.2, leading to slower training. As $d_s$ increases, efficiency improves while performance decreases. In practice, values around $d_s\approx 0.01$ provide a good trade-off. In Table~\ref{tab:hyper_token}, we show results adding token-level verification under different $d_t$. Setting $d_t=\infty$ equals sequence-level verification only. Enabling token-level verification significantly improves both accuracy and efficiency. Moreover, the results are stable across a reasonably wide range of $d_t$, indicating low sensitivity to choices. We use $d_t=2$ in the main experiments because this setting provides the best overall trade-off.

\subsection{Analysis Experiments}

\begin{table}[t]
\centering
\small
\setlength{\tabcolsep}{4pt}
\renewcommand{\arraystretch}{1.15}
\setlength{\abovecaptionskip}{2pt}
\resizebox{0.99\textwidth}{!}{
\begin{tabular}{l|cc|ccccccc}
\toprule
\textbf{Setting} & \textit{Training?} & \textit{Rollout?} & \textbf{AIME24} & \textbf{AIME25} & \textbf{AMC23} & \textbf{MATH500} & \textbf{AVG} &\textbf{OPS}& \textbf{GPU Days} \\
\midrule

GRPO~\cite{grpo} &  &  &40.3 & 29.5 & 81.4 & 78.0 & 57.3  & 0 & 98.1  \\
Partial~\cite{team2025kimi} &  &  & 23.7 & 19.4 & 67.5 & 73.2 & 45.9  &0.017& 51.6 \\
\midrule
Staleness-1    &  & \checkmark & 28.2 & 22.8 & 71.9 & 75.6 & 49.6 & 0.015 & 64.5 \\
GSPO~\cite{zheng2025gspo}          & \checkmark &  &  37.3 & 27.0 & 79.8 & 76.5 & 55.2 & 0.007 & 51.8 \\
SAPO~\cite{gao2025sapo}         & \checkmark &  & 37.5 & 27.7 & 79.6 & 75.7 & 55.1 & 0.008& \textbf{51.5} \\
VESPO~\cite{shen2026vespo}     & \checkmark &  & 31.0 & 26.2 & 74.3 & 74.5 & 51.5 & 0.010& 52.0 \\
\rowcolor{myBlue!60}
\textbf{RollVerify}    &  & \checkmark & \textbf{39.8} & \textbf{30.5} & \textbf{81.5} & \textbf{77.2} & \textbf{57.2} & \textbf{0.004} & 57.5 \\
\bottomrule
\end{tabular}
}
\caption{Comparison with other off-policy handling methods on Qwen3-8B-Base. GRPO is a fully on-policy baseline, and other methods use partial rollout. \textit{Training? }indicates that methods operate on loss level in training phase. \textit{Rollout?} indicates that methods operate on rollout samples.}
\label{tab:offpolicy_comparison}
\end{table}


\noindent \textbf{Comparison with Other Off-Policy Handling Methods.} We compare RollVerify with several methods that mitigate off-policy effects, including GSPO~\cite{zheng2025gspo}, SAPO~\cite{gao2025sapo}, and VESPO~\cite{shen2026vespo}. These methods modify the PPO-style objective during training, but do not change the rollout sample itself. As shown in Table~\ref{tab:offpolicy_comparison}, RollVerify outperforms naive partial rollout, matching GRPO while using fewer GPU days. It also achieves the lowest OPS among the methods using partial rollout. These results suggest that controlling the quality of rollout samples is more effective than loss reshaping.


\noindent \textbf{Different Context Lengths.} We evaluate RollVerify under maximum context lengths of 8K, 16K, 32K, and 64K. As reported in Table~\ref{tab:max_length_results}, longer contexts intensify the long-tail effect and thus lower overall efficiency. Nevertheless, RollVerify maintains model accuracy across all settings while delivering increasingly substantial speedups as the context length grows. The gain is relatively small at 8K, reducing training cost from 16.6 to 15.1 GPU days, but becomes much more significant at 64K, where the cost drops from 261 to 137 GPU days. These results suggest that RollVerify is especially beneficial for large-scale RL training scenarios involving long-context workloads.


\begin{table}[htbp]
\centering
\small
\setlength{\abovecaptionskip}{2pt}
\resizebox{0.99\textwidth}{!}{
\begin{tabular}{llcccccc}
\toprule
\textbf{Max Length} & \textbf{Setting} & \textbf{AIME24} & \textbf{AIME25}& \textbf{AMC23} & \textbf{MATH500} & \textbf{AVG} &  \textbf{GPU Days} \\
\midrule
\multirow{2}{*}{8K}
& on-policy  & 29.7   & 25.7   & 75.9   & 75.5   & 51.7   & 16.6~(1.0x)   \\
& RollVerify & 30.2   & 25.4   & 75.7   & 75.8   & 51.8   & 15.1~\textcolor{SoftGreen}{(1.1x)}   \\
\midrule
\multirow{2}{*}{16K}
& on-policy  & 36.1 & 28.8 & 80.6 & 72.6 & 54.5 & 51.8~(1.0x)\\
& RollVerify & 36.3 & 28.7 & 80.4 & 73.0 & 54.6 & 37.1~\textcolor{SoftGreen}{(1.4x)} \\
\midrule
\multirow{2}{*}{32K}
& on-policy  & 40.3 & 29.5 & 81.4 & 78.0 & 57.3 & 98.1~(1.0x)\\
& RollVerify & 39.8 & 30.5 & 81.5 & 77.2 & 57.2 & 57.5~\textcolor{SoftGreen}{(1.7x)} \\
\midrule
\multirow{2}{*}{64K}
& on-policy  & 48.0   & 36.0   & 88.1   & 78.7   & 62.7   & 261~(1.0x)   \\
& RollVerify & 47.9   & 36.5   & 88.5   & 78.5   & 62.8   & 137~\textcolor{SoftGreen}{(1.9x)}  \\
\bottomrule
\end{tabular}
}
\caption{Performance comparison under different maximum lengths.}
\label{tab:max_length_results}
\end{table}


\noindent \textbf{Verification Overhead}. The overhead introduced by RollVerify is small in practice. The main additional cost comes from OPS computation in Algorithm \ref{alg:OPS_compute}, which requires computing probabilities and is therefore roughly 2× the cost of ComputeProb. And the two-stage verification step is lightweight shown in Algorithm \ref{alg:two_stage_verify}, since both sequence-level and token-level verification only involve linear scans over the generated tokens, resulting in O(N) complexity. As shown in Table~\ref{tab:overhead}, the overall verification overhead remains low for both models, accounting for only a small fraction of the total training time. These results indicate that RollVerify adds only modest overhead while providing meaningful improvements in training sample quality.


\begin{table}[htbp]
\centering
\setlength{\abovecaptionskip}{2pt}
\resizebox{0.99\textwidth}{!}{
\begin{tabular}{c|c|cc|c}
\toprule
\multirow{2}{*}{Model} & \multirow{2}{*}{Setting} & \multicolumn{2}{c|}{Verification overhead} & \multirow{2}{*}{Step Time} \\
\cmidrule(lr){3-4}
 &  & OPS compute & Two-stage Verify &  \\
\midrule
Qwen3-8B-Base & RollVerify  & 20s (\textbf{6.4\%}) & 2s (\textbf{0.6\%}) & 310s \\
\midrule
Qwen3-30B-A3B-Base & RollVerify  & 25s (\textbf{6.9\%}) & 2s (\textbf{0.5\%})  & 362s \\
\bottomrule
\end{tabular}
}
\caption{Overhead under different settings. Step time indicates the average training step time.}
\label{tab:overhead}
\end{table}

\noindent \textbf{Generalization Experiments.} We further evaluate RollVerify on mathematical reasoning with DeepScaleR~\cite{deepscaler2025}, tool-assisted mathematical reasoning with ReTool~\cite{feng2025retoolreinforcementlearningstrategic}, and code generation~\cite{jain2024livecodebenchholisticcontaminationfree,chen2021codex}.
RollVerify outperforms partial rollout and achieves accuracy comparable to on-policy training in these settings. The code-generation results provide preliminary evidence beyond mathematics. More details are shown in Appendix~\ref{appendix:generalization_exp}.


\section{Conclusion and Limitations}
We present \textbf{RollVerify}, a verification-based framework for long-tail rollout RL. Built on partial rollout, it improves training sample quality through sequence-level and token-level verification, preserving on-policy performance while improving training efficiency. Experiments across different models, tasks, and context lengths demonstrate its effectiveness and robustness. Broader evaluation on non-math and agentic tasks, as well as validation in fully asynchronous RL training, remains future work.

\section*{Acknowledgements}
This work was supported by the National Key Research and Development Program of China (Grant No. 2023YFB4405102), the Postdoctoral Fellowship Program and the China Postdoctoral Science Foundation (Grant No. BX20250487), and in part by the Natural Science Foundation of Hunan Province (Grant No. 2024JJ6525).

\bibliographystyle{plain}
\bibliography{main}

\newpage

\appendix

\begin{table}[t]
\centering
\small
\setlength{\tabcolsep}{4pt}
\renewcommand{\arraystretch}{1.15}
\resizebox{0.95\textwidth}{!}{
\begin{tabular}{lccccccc}
\toprule
\textbf{Setting} & \textbf{Reward} & \textbf{Entropy}
& \textbf{AC Rate}
& \shortstack{\textbf{Lengths}}
& \textbf{OPS} & \textbf{AIME24/25} & \textbf{GPU Days} \\
\midrule
on-policy
& 0.665 & 0.113 & -- & 7118 & 0
& 38.7 / 32.8 & 81.3 (1.00$\times$) \\
RollVerify w/o switching
& 0.659 & 0.122 & 0.740 & 7109 & 0.005
& 38.7 / 32.5 & 64.3 (1.26$\times$) \\
RollVerify w/ switching
& 0.662 & 0.121 & 0.826 & 7201 & 0.005
& 38.8 / 32.8 & 60.2 (1.35$\times$) \\
\bottomrule
\end{tabular}
}
\caption{Effect of conditional switching under the 16K setting
with 700 training steps. RollVerify with switching switches to
on-policy rollout at step 598. AC Rate denotes the average
acceptance rate over training. Lengths mean response lengths. Speedups in parentheses are
relative to on-policy training.}
\label{tab:no_switching_700}
\end{table}

\section{Effect of Conditional Switching.}
\label{appendix:no_swtiching}
Conditional switching is designed to improve efficiency when the acceptance rate becomes low. To examine whether verification remains effective without switching back to on-policy rollout, we extend training to 700 steps under a maximum context length of 16K. We compare three settings: on-policy training, RollVerify without switching, and RollVerify with conditional switching. In the last setting, switching is triggered at step 598.

Table~\ref{tab:no_switching_700} reports the results. At step 700, RollVerify without switching achieves a reward of 0.659 and an entropy of 0.122, close to the on-policy baseline's 0.665 and 0.113, respectively. Its AIME24/25 scores are also comparable to the on-policy baseline (38.7/32.5 versus 38.7/32.8), while training cost decreases from 81.3 to 64.3 GPU days. Enabling conditional switching further reduces the cost to 60.2 GPU days, with AIME24/25 scores of 38.8/32.8. These endpoint results support the effectiveness of verification without switching over the evaluated horizon, while conditional switching provides additional efficiency gains.

\section{Generalization Experiments.}

\label{appendix:generalization_exp}

We conduct additional experiments on mathematical reasoning, tool-assisted mathematical reasoning, and code generation. For mathematical reasoning, we use the DeepScaleR dataset and Qwen3-8B-Base. For tool-assisted mathematical reasoning, we follow the original ReTool setting and use a model obtained by supervised fine-tuning of Qwen2.5-7B-Instruct~\cite{qwen2025qwen25technicalreport}. For code generation, we use training data from DeepCoder~\cite{deepcoder2025} and Qwen3-8B-Base as the base model. As shown in Tables~\ref{tab:dataset_results} and Table~\ref{tab:coding_results}, RollVerify achieves higher accuracy than partial rollout and accuracy comparable to on-policy training, while reducing training cost relative to on-policy training. The code-generation results provide preliminary evidence beyond mathematics.



\begin{table}[htbp]
\centering
\small
\setlength{\abovecaptionskip}{2pt}
\begin{tabular}{llcccccc}
\toprule
\textbf{Dataset} & \textbf{Setting} & \textbf{AIME24} & \textbf{AIME25 }& \textbf{AMC23} & \textbf{MATH500} & \textbf{AVG} & \textbf{GPU Days} \\
\midrule
\multirow{3}{*}{DeepScaleR}
& on-policy        & 38.9 & 28.4 & 78.9 & 89.8 & 59.0 & 91.2 (1.0x)    \\
& partial & 25.1 & 20.7 & 67.9 & 84.4 & 49.5 & 47.2 (1.9x) \\
\rowcolor{myBlue!60}
& RollVerify       & 38.8 & 28.5 & 79.4 & 90.1 & 59.2 & 52.1~\textcolor{SoftGreen}{(1.8x)} \\
\midrule
\multirow{3}{*}{ReTool}
& on-policy  & 39.4   & 32.0   & 74.0   & 68.9   & 53.6   & 102 (1.0x)           \\
& partial &  28.9  & 20.0   & 69.0   & 60.8   & 44.7   & 50.5    (2.0x)        \\
\rowcolor{myBlue!60}
& RollVerify       & 39.9   & 31.7   & 73.9   & 68.5   & 53.5   & 63.5~\textcolor{SoftGreen}{(1.6x)}            \\
\bottomrule
\end{tabular}
\caption{Performance comparison across datasets and settings.}
\label{tab:dataset_results}
\end{table}


\begin{table}[htbp]
\centering
\small
\setlength{\abovecaptionskip}{2pt}
\begin{tabular}{lcccc}
\toprule
\textbf{Setting} & \textbf{LiveCodeBench} & \textbf{HumanEval} & \textbf{AVG} & \textbf{GPU Days} \\
\midrule
on-policy       & 28.1& 85.2 & 56.6 & 37.0~(1.0x) \\
partial & 24.7 & 80.3 & 52.5 & 24.7~(1.5x) \\
RollVerify       & 27.9 & 85.1 & 56.5 & 26.4~(1.4x) \\
\bottomrule
\end{tabular}
\caption{Performance comparison on coding benchmarks.}
\label{tab:coding_results}
\end{table}






\newpage
\section*{NeurIPS Paper Checklist}

\begin{enumerate}

\item {\bf Claims}
    \item[] Question: Do the main claims made in the abstract and introduction accurately reflect the paper's contributions and scope?
    \item[] Answer: \answerYes{} 
    \item[] Justification: The main claims in the abstract and introduction accurately reflect the paper’s contributions and scope.

\item {\bf Limitations}
    \item[] Question: Does the paper discuss the limitations of the work performed by the authors?
    \item[] Answer: \answerYes{} 
    \item[] Justification: The paper provides a brief discussion of the limitations in the Conclusion section.

\item {\bf Theory assumptions and proofs}
    \item[] Question: For each theoretical result, does the paper provide the full set of assumptions and a complete (and correct) proof?
    \item[] Answer: \answerNA{} 
    \item[] Justification: The paper is primarily based on empirical observations and proposes heuristic approaches, which are not fully supported by formal mathematical proofs or a complete set of theoretical assumptions.

    \item {\bf Experimental result reproducibility}
    \item[] Question: Does the paper fully disclose all the information needed to reproduce the main experimental results of the paper to the extent that it affects the main claims and/or conclusions of the paper (regardless of whether the code and data are provided or not)?
    \item[] Answer: \answerYes{} 
    \item[] Justification: The experimental results can be reproduced based on the information provided in the paper, and the authors have indicated that the code will be released in the future.

\item {\bf Open access to data and code}
    \item[] Question: Does the paper provide open access to the data and code, with sufficient instructions to faithfully reproduce the main experimental results, as described in supplemental material?
    \item[] Answer: \answerNo{} 
    \item[] Justification: The code and data are not yet publicly available, and although the authors plan to release them later, they are currently not provided with sufficient instructions in the supplemental material to ensure faithful reproduction of the results.

\item {\bf Experimental setting/details}
    \item[] Question: Does the paper specify all the training and test details (e.g., data splits, hyperparameters, how they were chosen, type of optimizer) necessary to understand the results?
    \item[] Answer: \answerYes{} 
    \item[] Justification: The full details are provided in the main text and supplemental material.

\item {\bf Experiment statistical significance}
    \item[] Question: Does the paper report error bars suitably and correctly defined or other appropriate information about the statistical significance of the experiments?
    \item[] Answer: \answerNo{} 
    \item[] Justification: We report benchmark accuracies and training costs, but do not provide error bars, confidence intervals, or statistical significance tests.


\item {\bf Experiments compute resources}
    \item[] Question: For each experiment, does the paper provide sufficient information on the computer resources (type of compute workers, memory, time of execution) needed to reproduce the experiments?
    \item[] Answer: \answerYes{} 
    \item[] Justification: The paper provides sufficient information about the computational resources used, including details in both the main text and the Appendix.
    
\item {\bf Code of ethics}
    \item[] Question: Does the research conducted in the paper conform, in every respect, with the NeurIPS Code of Ethics \url{https://neurips.cc/public/EthicsGuidelines}?
    \item[] Answer: \answerYes{} 
    \item[] Justification: The research presented in the paper fully conforms to the NeurIPS Code of Ethics in all respects.

\item {\bf Broader impacts}
    \item[] Question: Does the paper discuss both potential positive societal impacts and negative societal impacts of the work performed?
    \item[] Answer: \answerNA{} 
    \item[] Justification: There is no societal impact of the work performed.
    
\item {\bf Safeguards}
    \item[] Question: Does the paper describe safeguards that have been put in place for responsible release of data or models that have a high risk for misuse (e.g., pre-trained language models, image generators, or scraped datasets)?
    \item[] Answer: \answerNA{} 
    \item[] Justification: The paper poses no such risks.

\item {\bf Licenses for existing assets}
    \item[] Question: Are the creators or original owners of assets (e.g., code, data, models), used in the paper, properly credited and are the license and terms of use explicitly mentioned and properly respected?
    \item[] Answer: \answerYes{} 
    \item[] Justification: All resources used in the paper comply with their respective licenses and terms of use.

\item {\bf New assets}
    \item[] Question: Are new assets introduced in the paper well documented and is the documentation provided alongside the assets?
    \item[] Answer: \answerNA{} 
    \item[] Justification: The paper does not release new assets.

\item {\bf Crowdsourcing and research with human subjects}
    \item[] Question: For crowdsourcing experiments and research with human subjects, does the paper include the full text of instructions given to participants and screenshots, if applicable, as well as details about compensation (if any)? 
    \item[] Answer: \answerNA{} 
    \item[] Justification: The paper does not involve crowdsourcing nor research with human subjects.

\item {\bf Institutional review board (IRB) approvals or equivalent for research with human subjects}
    \item[] Question: Does the paper describe potential risks incurred by study participants, whether such risks were disclosed to the subjects, and whether Institutional Review Board (IRB) approvals (or an equivalent approval/review based on the requirements of your country or institution) were obtained?
    \item[] Answer: \answerNA{} 
    \item[] Justification: The paper does not involve crowdsourcing nor research with human subjects.

\item {\bf Declaration of LLM usage}
    \item[] Question: Does the paper describe the usage of LLMs if it is an important, original, or non-standard component of the core methods in this research? Note that if the LLM is used only for writing, editing, or formatting purposes and does \emph{not} impact the core methodology, scientific rigor, or originality of the research, declaration is not required.
    \item[] Answer: \answerNA{} 
    \item[] Justification: The core method development in this research does not involve LLMs as any important, original, or non-standard components.

\end{enumerate}

\end{document}